\documentclass[letterpaper, 10 pt, conference]{ieeeconf}  

\IEEEoverridecommandlockouts                              
\usepackage{graphicx} 
\usepackage{caption}
\usepackage{subcaption}
\usepackage{amsmath}
\usepackage{amsfonts}
\usepackage[cal=rsfso,calscaled=.96]{mathalfa}
\usepackage{mathrsfs}  
\usepackage{txfonts} 
\usepackage{enumerate}
 
\usepackage{hyperref}  
\usepackage{amsmath}
\usepackage{algorithm}
\usepackage[noend]{algpseudocode}

\usepackage{times} 
\usepackage{amssymb,longtable,calc}
\usepackage{mathptmx}       
\usepackage{graphicx}
\usepackage[textwidth=2cm,colorinlistoftodos]{todonotes}
\usepackage{xspace}
\usepackage{latexsym}
\usepackage{amsmath,amssymb,graphicx,xspace}
\usepackage{times}   
\usepackage{multirow}
\usepackage{color,comment,rotating,url,xspace} 
\usepackage{tikz}
\usetikzlibrary{arrows,shadows,shapes,backgrounds,decorations,snakes,fit}
\usepackage{sidecap}
\usepackage{amsmath}

\title{\LARGE \bf
 Anticipating many futures: Online human motion prediction and synthesis for human-robot collaboration
}

\author{Judith B\"utepage, Hedvig Kjellstr\"om and Danica Kragic 
\thanks{Authors are with the Department of Robotics, Perception and Learning, KTH Royal Institute of Technology, Sweden.
        {\tt\small butepage|hedvig|dani@kth.se}}%
}

\begin{document}

\maketitle
\thispagestyle{empty}
\pagestyle{empty}

\begin{abstract}
 Fluent and safe interactions of humans and robots require both partners to anticipate the others' actions. A common approach to human intention inference is to model specific trajectories towards known goals with supervised classifiers. However, these approaches do not take possible future movements into account nor do they make use of
 kinematic cues, such as legible and predictable motion. The bottleneck of these methods is the lack of an accurate model of general human motion. In this work, we present a conditional variational autoencoder that is trained to predict a window of future human motion given a window of past frames. Using skeletal data obtained from RGB depth images, we show how this unsupervised approach can be used for online motion prediction for up to 1660 ms. Additionally, we demonstrate online target prediction within the first 300-500 ms after motion onset without the use of target specific training data. The advantage of our probabilistic approach is the possibility to draw samples of possible future motions. Finally, we investigate how movements and kinematic cues are represented on the learned low dimensional manifold.   
\end{abstract}
\section{INTRODUCTION}
\label{sec:intro}

Close interaction and collaboration with humans in shared workspaces requires robots to be able to coordinate their actions in space and time with their human partners. The robot's actions need to be interpretable for the human partner to make the interaction intuitive.
Similarly, human actions need to be identified on the macro level, e.g. grasping an object, and on the micro level, e.g. the exact trajectory and velocity of the movement. Online analysis of the current human motion is not sufficient to guarantee  a fluent and safe interaction. Instead, the robot needs to be able to anticipate human motion and actions to be able to initiate actions in time and to render long--term planning possible. 

In general, the topic of prediction in Human--Robot Interaction (HRI) can be posed as two research questions:
\begin{enumerate}
\item How can the robot anticipate and understand human actions in order to facilitate decision making in a joint setting? 
\item How can the robot plan its actions such that they are intuitive and readable to a human observer?
\end{enumerate}

A number of studies have addressed question 1) by focusing on target prediction of human reaching movements \cite{perez2015fast, mainprice2013human,maeda2016anticipative,luo2015framework}. This problem is often tackled with supervised trajectory classification. However, in most cases, these approaches only incorporate knowledge about the current arm configurations and do not anticipate human motion dynamics. Additionally, they depend on the training data and cannot generalize to novel targets.
\begin{figure}[t!]
    \includegraphics[width=1\linewidth]{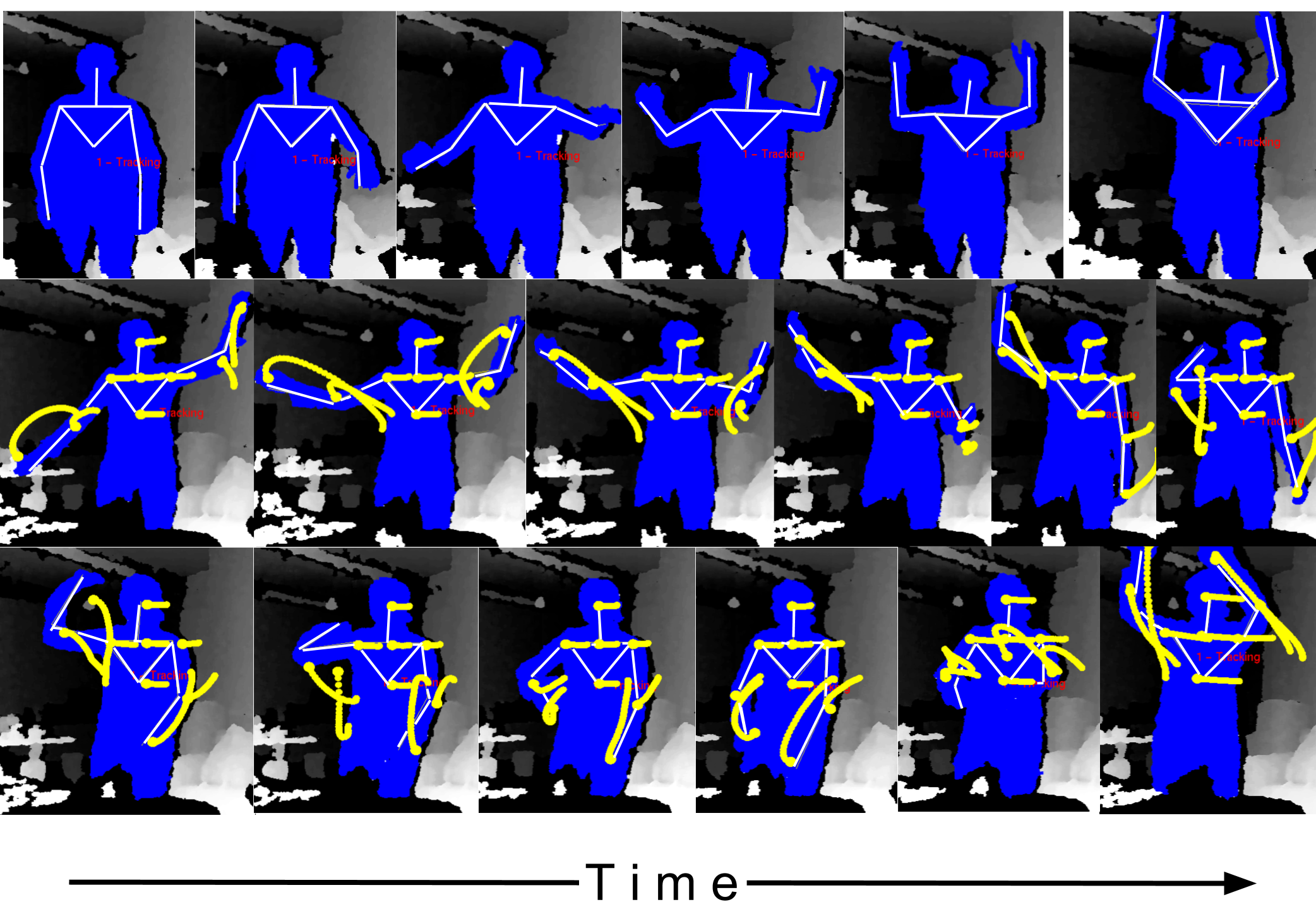} 
    \caption{ Top row: As an example of training data we depict the depth image with the tracked upper body skeleton.
    Bottom rows: Given a window of past frames, we show the samples of future joint trajectories as yellow dots. These samples cover approximately 1660 ms of future motion. Each row depicts approximately 2 seconds of movements.} \label{fig:front} \vspace{-0.3cm}
\end{figure}

The bottleneck of an anticipatory approach is an accurate model of general human motion. In order to make inferences about future movements, such a model is required to be able to distinguish between subtle kinematic cues about intentions which humans send during interaction \cite{sartori2009does, sacheli2013kinematics,  scorolli2014give}. 

This idea closely relates to the concepts of legibility and predictability of motion trajectories \cite{dragan2013legibility}. A legible motion trajectory renders the inference of the goal given a partial trajectory observation possible. A predictable motion trajectory is defined as the path one would expect given a known goal. While these concepts have been applied to robot motion planning \cite{dragan2014integrating}, i.e. addressing  question 2), they have not been studied in the context of human motion. Additionally, the top--down approach introduced in \cite{dragan2014integrating} does not rely on prediction but transforms a cost function into a probability distribution which requires computationally expensive approximations of integrals. 

Following concepts from neuroscience and psychology we address question 1) with a sensorimotor, bottom--up approach of internal forward models for action simulation and understanding in a social context \cite{pezzulo2011should}. Forward models in the common sense describe the sensory change induced by one's own actions. In a social context, these models can be employed to predict the outcome of  others' actions, while treating the motor parameters of these actions as hidden variables that need to be approximated. As depicted in Fig. \ref{fig:front}, our approach allows to generate possible future trajectories of human motion given past observations. By focusing on generative models, we believe that a robotic system which integrates action generation and action understanding on a representational and modeling level will facilitate human--robot collaboration and constitute a bridge between question 1) and 2).  
 
\subsection{Contributions and paper outline}

In this work we extend our previous work \cite{butepage17} and propose a model that is based on recent advances in deep generative models, namely variational autoencoders (VAE) \cite{kingma2013auto}. By transforming the VAE into a conditional variational  autoencoder (CVAE) \cite{sohn2015learning}, we are able to predict human skeletal motion online for up to 1660 ms from RGB depth images such as depicted in  the top row of Fig. \ref{fig:front}. This model is well suited for predicting human motion as it can capture dynamic, non--linear dependencies under uncertainty. Furthermore, the Bayesian approach allows us to approximate the posterior distribution over hidden variables and a predictive likelihood of trajectories under noisy observations. Thus, the CVAE provides us both with a measure of uncertainty over predictions and a possibility to draw samples of future motions given the current observed motion trajectory. The two bottom rows of Fig. \ref{fig:front} depict such samples of possible upper body movements during online recordings. We demonstrate how these inferences can be used to predict and sample general future human motion. Additionally, we show that the predictions made by our model can be used to determine the goal of reaching actions without the use of target specific training data. By investigating the learned representation, we find that legible and predictable motions are automatically clustered by the model.
The contribution of this work is a biologically inspired probabilistic forward model of human motion that can be employed in online human--robot collaboration settings. 

This paper is organized as follows. We begin the method section by presenting the  data representation in Sec. \ref{subsec:method_data}. As our model is based on variational autoencoders, we summarize the concepts behind VAEs in Sec. \ref{subsec:method_VAE} before describing our temporal CVAE in Sec. \ref{subsec:method_VTE}. Finally, we exemplify how the inferences made with help of the CVAE can be projected back into the context of the environment on a target prediction task in Sec. \ref{subsec:method_goal}. In the experiments, we shortly describe the data collection and model specifications in Sec. \ref{sec:models}. Subsequently, we present the predictive performance of our approach in Sec. \ref{sec:onlinepred} and the performance in end-point target classification in Sec. \ref{sec:target}. Finally, we demonstrate the advantage of our probabilistic representation learning method by sampling future human motions in Sec. \ref{sec:sample} and by investigating how legible and predictable movements are represented on the learned low dimensional manifold in Sec. \ref{sec:disent}. We conclude the paper in Sec. \ref{sec:conc} and outline future work.

\section{RELATED WORK}
\label{sec:related}

Prediction of human activities in a collaborative HRI setting has been addressed by several groups. While the computer vision community has addressed the topic of skeletal human motion prediction with neural networks in a variety of works \cite{taylor2006modeling, fragkiadaki2015recurrent,jain2015structural} in this section we will focus on related work with a HRI context. As our approach is trajectory based, we will also discard high--level approaches towards collaborative HRI such as affordance based coordination \cite{koppula2016anticipatory}.  

Among systems concerned with low--level dynamics, we can distinguish between supervised and unsupervised approaches. 

Supervised approaches commonly use information about the trajectory type, such as end-goal position, in order to estimate task specific models for prediction and classification. To account for differences in trajectory execution, in \cite{perez2015fast} the trajectories are aligned with help of dynamic time warping (DTW) before fitting multivariate Gaussian distributions to each time step. Task dependency is introduced by modeling the conditional likelihood of a trajectory given a  motion class and task. With help of this method, the goal of a reaching motion can be classified with 70 \% accuracy within 500 ms after motion onset.  

In addition to end--point prediction, the position of the human with respect to the robot needs to be taken into account. In order to estimate the area of the task space that will be occupied by the human during a goal--directed motion, Mainprice et al. \cite{mainprice2013human} rely on Gaussian Mixture Models (GMM) and Gaussian Mixture Regression (GMR) to estimate trajectory and occupancy models of the human motion. Subsequently, these estimates are taken into account by the robot during action planning. 

In a collaborative setting, the robot  needs to consider the state of the human partner with respect to itself within the task space. In \cite{amor2014interaction}, interaction primitives are employed to anticipate human actions within task context. Interaction primitives \cite{maeda2016anticipative} are dynamic motor primitives that model both the robot's and human's dynamics as well as the correlation between their actions in a specific task. By utilizing look-up tables of possible interaction sequences and by conditioning on the recent trajectories, \cite{amor2014interaction} shows how the next human action can be anticipated and integrated into the robot's action selection process.

Unsupervised methods can be used to learn the dynamics of motion without explicitly defining high--level parameters that restrict the learning of structure. 
To accomplish this, in \cite{luo2015framework} GMMs and GMR are combined with a clustering algorithm to both learn the dynamics of the motions and to identify trajectories. 

In comparison to the majority of these approaches, we do not rely on a model of the current state of the human body but anticipate future trajectories. This allows us to infer an intended target without specific training data. As the evaluation of a neural network requires only feedforward matrix multiplications, it can be used in an online setting. The Bayesian approach is able to incorporate the noisy data received from the RGB depth images and to operate on a low frame rate. Except for \cite{mainprice2013human}, the approaches presented above \cite{perez2015fast,maeda2016anticipative, luo2015framework} all rely on motion capture systems with higher frame rates.  

Finally, we include a work that is not concerned with 3D trajectory prediction but with general motion prediction in images as it is related by methodology and argumentation. Walker et al. \cite{walker2016uncertain} present a conditional variational autoencoder to predict the trajectory of all pixels in an image over a given time span. Given a certain image, the model can be used to generate future trajectories of all pixels. Being unsupervised, the information about e.g. human poses is not explicitly represented by the model. Since this approach is computationally exhaustive and does not operate in 3D, it is limited in its applicability to online HRI. 

\section{METHOD}
\label{sec:method}

In this section we introduce a deep generative model for human motion prediction and synthesis that can be employed in online HRI tasks. The stochastic computational graph of this model is presented  on the left in Fig. \ref{fig:generativemodel}.
The main idea of our approach is to predict a window of frames at future times steps, given the last time steps. For this, we model a joint distribution over these two variables and a number of hidden variables which govern the unobserved dynamics. In Fig. \ref{fig:generativemodel} the observed variables at time $t-1$ and time $t$ are given by $\mathbf{x_{t-1}}$ and $\mathbf{x_{t}}$, while the hidden variables are represented by $\mathbf{z_{t-1}}$ and $\mathbf{z_{t}}$. In order to allow for expressive models, we employ conditional variational autoencoders in which the encoder $h_e$, the inner layers $h_t$ and the decoder $h_d$ are represented by fully--connected neural networks with feedforward layers as shown on the right of Fig. \ref{fig:generativemodel}.

\subsection{Data representation}
\label{subsec:method_data}

We represent the human upper body as a vector of Cartesian coordinates $x,y,z$ for each joint.  Given a number of $N_j$ joints and a time step $t$, this results in a matrix $\mathbf{f}_t = [f_{t,i,x}, f_{t,i,y}, f_{t,i,z}]_{i=1:N_j}$ of dimension $(N_j,3)$. A time window covering the interval $[t:t+\Delta t]$ is given by the concatenation of frames $\mathbf{F}_{t:(t+\Delta t)} = [\mathbf{f}_t, \mathbf{f}_{t+1}, ... \mathbf{f}_{t+\Delta t}]$ of dimension $(\Delta t,N_j,3)$. For simplicity,  the past $\Delta t$ frames are denoted by $\mathbf{F}_{t-\Delta t} := \mathbf{F}_{(t-\Delta t):t}$ and the future $\Delta t$ frames are denoted by $\mathbf{F}_{t+\Delta t} := \mathbf{F}_{(t+1):(t+1+\Delta t)}$. 
Thus, the past observation $\mathbf{x}_{t-1}$ as shown in Fig. \ref{fig:generativemodel} is given by a vectorization of the matrix $\mathbf{F}_{t-\Delta t}$ and the next observation $\mathbf{x}_{t}$ is given by a vectorization of the matrix $\mathbf{F}_{t+\Delta t}$. We will discuss the CVAE in terms of the more general  $\mathbf{x}_{t-1}$ and  $\mathbf{x}_{t}$ while using $\mathbf{F}_{t-\Delta t}$ and $\mathbf{F}_{t+\Delta t}$ in application specific sections. 

\subsection{Background of variational autoencoders}
\label{subsec:method_VAE}

Generative models are characterized by a joint distribution $p_\theta(\mathbf{x}, \mathbf{z})$ over observed random variables $\mathbf{x}$ and hidden random variables $\mathbf{z}$ which can be characterized by parameters $\theta$. We denote a distribution $p$ that is parameterized by $\theta$ by $p_\theta$ and the $i$th observation of a random variable $\mathbf{x}$ by $\mathbf{x}_i$.  

In order to make inferences over the hidden variables and the parameters, we require a tractable posterior $p_\theta(\mathbf{z} | \mathbf{x})$ for any instantiation of $\mathbf{z}$ and $\mathbf{x}$ which can be acquired through Bayes' theorem. However, the model evidence $p_\theta(\mathbf{x}) = \int p_\theta(\mathbf{z})p_\theta(\mathbf{x}|\mathbf{z})d\mathbf{z}$ is often intractable. 
Variational autoencoders \cite{kingma2013auto} have been proposed to circumvent this intractability by approximating the true posterior  $p_\theta(\mathbf{z} | \mathbf{x})$ with help of a probabilistic recognition model, the encoder, $q_\phi(\mathbf{z} | \mathbf{x})$ with parameters $\phi$. Likewise, the data likelihood  $p_\theta(\mathbf{x} | \mathbf{z})$ can be modeled by a probabilistic decoder model. It can be shown that the  marginal likelihood of sample $\mathbf{x}_i$ can be approximated by maximizing the lower bound 
\begin{equation}
\mathcal{L}(\phi, \theta, \mathbf{x}_i) = -D_{KL}(q_i||p_\theta(\mathbf{z})) 
+ \mathbb{E}_{q_i}[log(p_\theta(\mathbf{x}_i|\mathbf{z}))],
\end{equation}
where $q_i$ denotes $q_\phi(\mathbf{z} | \mathbf{x}_i)$, $D_{KL}(q||p)$ is the Kullback--Leibler divergence between the distributions $q$ and $p$ and $\mathbb{E}_{p_\theta} [f(\mathbf{x})]$ is the expectation of a function $f(\mathbf{x})$ under the distribution $p_\theta(\mathbf{x})$. 
Variational autoencoders model $q_\phi(\mathbf{z} | \mathbf{x}_i)$ with help of a differentiable transformation $g_\phi$ such that $\mathbf{z} \sim g_\phi( \mathbf{\epsilon}, \mathbf{x}_i )$ where $\mathbf{\epsilon}$ is a noise variable $\mathbf{\epsilon} \sim p(\mathbf{\epsilon})$. Using ideas from  Stochastic
Gradient Variational Bayes, the lower bound can then be approximated by averaging over $S$ samples as follows
\begin{equation}
\hat{\mathcal{L}}(\phi, \theta, \mathbf{x}_i) = \frac{1}{S}\sum_{s=1}^S  log(p_\theta(\mathbf{x}_i,\mathbf{z}_i^s)) - log(q_\phi(\mathbf{z}_i^s| \mathbf{x}_i)),
\end{equation} 
where the hidden variables are given by $\mathbf{z}_i^s =g_\phi( \mathbf{\epsilon}_s, \mathbf{x}_i )$ and $ \mathbf{\epsilon}_s\sim p(\mathbf{\epsilon})$.
Modeling the encoder and decoder with neural networks allows for an expressive model that can describe non--linear mappings between the hidden and the observed variables while leveraging the advantages of the Bayesian approach. With help of the so called reparameterization trick the parameters $\theta$ and $\phi$ can be efficiently optimized by backpropagation.  

\tikzstyle{unobserved} = [draw, circle, node distance=1.5cm,text centered, text width=1.5em]
\tikzstyle{observed} = [draw, circle,fill=gray!40, node distance=1.5cm,text centered, text width=1.5em]
\begin{figure}
\vspace{0.2cm}
\centering   
\pgfdeclarelayer{background}
\pgfdeclarelayer{foreground}
\pgfsetlayers{background,main,foreground}

\begin{tikzpicture}

\tikzstyle{surround} = [thick,draw=black,rounded corners=1mm]
\tikzstyle{scalarnode} = [circle, draw, fill=white!11,  
    text width=1.4em, text badly centered, inner sep=2.5pt]
\tikzstyle{scalarquad} = [rectangle, draw, fill=white!11,  
    text width=1.4em, text badly centered, inner sep=2.5pt]
\tikzstyle{scalarquadlong} = [rectangle, draw, fill=white!11,  
    text width=3em, text badly centered, inner sep=2.5pt]  
 
 \tikzstyle{scalarquadtext} = [rectangle,  fill=white!11,  
    text width=1.4em, text badly centered, inner sep=2.5pt]
\tikzstyle{arrowline} = [draw,color=black, -latex]

	\node [scalarnode] at (-5,0)   (gzt1)  {$\mathbf{z}_{t-1}$};
	\node [scalarnode] at (-3.5,0) (gzt)   {$\mathbf{z}_{t}$};
    \node [scalarquad] at (-5,-1.5)   (ge)  {$h_e$};
	\node [scalarquad] at (-3.5,-1.5) (gt)   {$h_t$};
    \node [scalarquad] at (-2,-1.5) (gd)   {$h_d$};
    \node [scalarnode, fill=black!30 ] at (-5,-3)   (gxt1)  {$\mathbf{x}_{t-1}$};
	\node [scalarnode, fill=black!30 ] at (-2,-3) (gxt)   {$\mathbf{x}_{t}$};
    
    \path [arrowline] (gxt1) to (ge);
    \path [arrowline] (gzt1) to (ge);
    \path [arrowline] (gzt) to (gt);
    \path [arrowline] (ge) to (gt);
    \path [arrowline] (gt) to (gd);
    \path [arrowline] (gd) to (gxt);
    
    \node [scalarquadlong, fill=black!30] at (0,0.5) (l1)   {$\mathbf{x}_{t-1}$};
    \node [scalarquadlong] at (0,-0.2) (l2)   {200};
    \node [scalarquadlong] at (0,-0.9) (l3) {  20 };
     \node [scalarquadtext] at (1.3,-0.9) (l3b)   {$\mathbf{z}_{t-1}$};
    \node [scalarquadlong] at (0,-1.6) (l4)   {30 };
    \node [scalarquadlong] at (0,-2.3) (l5)   { 20 };
    \node [scalarquadtext] at (1.3,-2.3) (l5b)   {$\mathbf{z}_{t}$};
    \node [scalarquadlong] at (0,-3) (l6)   {200 };
    \node [scalarquadlong, fill=black!30] at (0,-3.7) (l7)   {$\mathbf{x}_{t}$};
    
    \path [arrowline] (l1) to (l2);
    \path [arrowline] (l2) to (l3);
    \path [arrowline] (l3) to (l4);
    \path [arrowline] (l4) to (l5);
    \path [arrowline] (l5) to (l6);
    \path [arrowline] (l6) to (l7);
    
    \path [arrowline] (l3b) to (l3);
    \path [arrowline] (l5b) to (l5);
    
    

\end{tikzpicture}
\caption{Left: Stochastic computational graph of the generative model CVAE. Gray circles represent observed random variables, white circles represent unobserved random variables and white squares represent parameterized mappings. Right: Feedforward network structure with the number of units in each layer.  } \label{fig:generativemodel} \vspace{-0.3cm}
\end{figure}
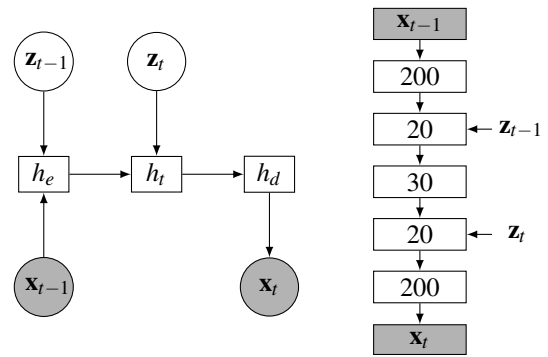

\subsection{Conditional variational autoencoders}
\label{subsec:method_VTE}

While VAEs are pure autoencoders, i.e. the likelihood of the data depends only on a single observed variable, we are interested in predicting future time steps given past observations. Thus, we require a sequential generative model that encodes the dynamics of the time series. In order to allow online evaluations of the model, we aim at a feedforward model under the markov assumption. We propose an approach similar to the ideas presented in \cite{sohn2015learning} and \cite{rezende2016one} but do not consider recurrent connections and extensions. 
Instead, we assume that the generative process consists of three components: an encoder $h_e$, a transitioner $h_t$ and a decoder $h_d$, depending on parameters $\theta^e,\theta^t$ and $\theta^d$ respectively. In general terms, the encoding and decoding of an observation $\mathbf{x}_{t-1}$ to predict the next observation $\mathbf{x}_{t}$ is described by
\begin{align}
\text{latent variables }& \mathbf{z}_t \sim \mathcal{N}(\mathbf{z}_t|\mathbf{0}, \mathbf{I})   \label {eq:lv}\\ 
\text{encoded }& \mathbf{e}_t = h_e(\mathbf{z}_{t-1}, \mathbf{x}_{t-1}, \theta^e) \label{eq:e}\\
\text{transitioned }& \mathbf{t}_t = h_t(\mathbf{z}_{t}, \mathbf{e}_t, \theta^t)  \label{eq:t}\\
\text{observation }& \hat{\mathbf{x}}_t \sim p_{\theta^d}(\mathbf{x}_t| h_d(\mathbf{t}_t, \theta^d)) \label{eq:o}.
\end{align}
At each time step $t$, the encoder maps the last observation $\mathbf{x}_{t-1}$ and a sample from the prior to an encoded state $\mathbf{e}_t$, see Equation \ref{eq:lv}. This encoded state is passed through the transitioner to encode the dynamics in the hidden variables. The resulting state $\mathbf{t}_t$ is decoded into the probability distribution over observations $\mathbf{x}_t$. A sample from this output distribution is denoted by $\hat{\mathbf{x}}_t$. Both the output of the encoder and the transitioner map the sample of the prior onto a non--linear feature space with help of a mean and covariance estimate. 

Note that the hidden sequence does only depend on the last observation and not the previous hidden states for $t' < t-1$. Thus, this model is not a hidden markov or recurrent model but assumes conditional independence $\mathbf{x}_t \Perp \mathbf{x}_{t'} | \mathbf{x}_{t-1} \ \forall t' < t-1 $. The encoder-decoder structure facilitates the learning of a low dimensional manifold of the data dynamics.
When fixing all parameters, the distribution of $\mathbf{x}_t$ can be viewed as a conditional distribution $p_{\lambda}(\mathbf{x}_t| \mathbf{x}_{t-1}, \mathbf{z}_{t-1:t})$, where  $\lambda  =  (\theta^e,\theta^t$, $\theta^d)$.
In order to model continuous outputs, we define the output of the decoder $p_{d}(\mathbf{x}_t| h_d(\mathbf{t}_t, \theta^d))$ to be normal distributed with a mean $\mu^d$ and diagonal covariance matrix $\Sigma^d $ which are  functions of $\mathbf{t}_t$ parameterized by $\theta^d$. As $p_{\theta^d}(\mathbf{x}_t| h_d(\mathbf{t}_t, \theta_d))$  treats the output dimensions independently, we can make predictions for a single joint $j$, by considering only the corresponding dimensions.

\subsection{Target inference of reaching motions}
\label{subsec:method_goal}

While our model is unsupervised it is applicable to target prediction. For this, we do not gather target specific training data but use the predictive power of our model to infer future positions. Given a number of $N_g$, equally likely targets, let $\boldsymbol{\mu}^g$ be the position of target $g \in [1,N_g]$ in Cartesian coordinates. Furthermore, let $\hat{\mathbf{f}}_{t,j} \sim \mathcal{N}(\mathbf{f}_{t,j}|\mu^d_{I_j}, \Sigma^d_{I_j})$ be the  predictive distribution of the $j$th joint at time $t$, determined by $p_\lambda(\mathbf{f}_{t,j}| \mathbf{F}_{t-\Delta t}, \mathbf{z}_{t-1:t})$. Here $I_j$ is the index set of all dimensions that correspond to joint $j$ and $\hat{\mathbf{f}}_{t,j}$ indicates a sample of the distribution.

We model the location of a target $g$ by a multivariate Gaussian distribution with mean $\boldsymbol{\mu}^g$ and isotropic  covariance function $\boldsymbol{\Sigma}^g$. Thus,  the likelihood of a single prediction $\hat{\mathbf{f}}_{t,j}$ at time $t$ of joint $j$ being aimed at goal $g$ is given by $p(\hat{\mathbf{f}}_{t,j}|g) \sim  \mathcal{N}(\hat{\mathbf{f}}_{t,j}|\boldsymbol{\mu}^g, \boldsymbol{\Sigma}^g)$.
The probability of a goal given an observation is therefore given by
\begin{align}
p(g|\hat{\mathbf{f}}_{t,j})&p_\lambda(\hat{\mathbf{f}}_{t,j}| \mathbf{F}_{t-\Delta t}, \mathbf{z}_{t-1:t}) \propto p( \hat{\mathbf{f}}_{t,j}|g)p_\lambda(\hat{\mathbf{f}}_{t,j}| \mathbf{F}_{t-\Delta t}, \mathbf{z}_{t-1:t})  \nonumber \\ 
&\propto \mathcal{N}(\hat{\mathbf{f}}_{t,j}|  \mu^g  , \Sigma^g) \mathcal{N}(\hat{\mathbf{f}}_{t,j}| \mu^d_{I_j}, \Sigma^{d}_{I_j}).
\end{align}
 
For an entire  sequence, we have
\begin{align}
p(g|\hat{\mathbf{F}}_{t+\Delta t,j})&p_{\theta^d}(\hat{\mathbf{F}}_{t+\Delta t,j}| \mathbf{F}_{t-\Delta t}, \mathbf{z}_{t-1:t}) \nonumber \\ 
&\propto \prod_{\tau=t+1}^{t+1+\Delta t} p(g|\hat{\mathbf{f}}_{\tau,j})p_{\theta^d}(\hat{\mathbf{f}}_{\tau,j}| \mathbf{F}_{t-\Delta t}, \mathbf{z}_{t-1:t}) \label{eq:goal}.
\end{align}
We compute the probability of target $g$ by normalizing over all targets.

\begin{figure}[b!]
    \includegraphics[width=0.9\linewidth]{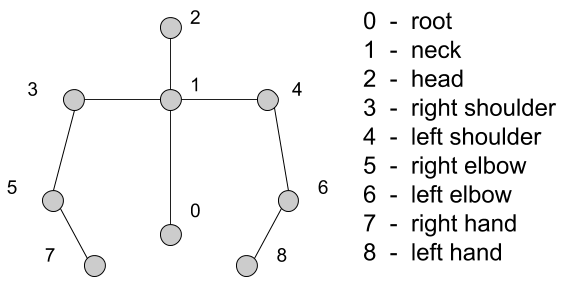} 
    \caption{The joints of the skeleton used in this work.
    } \label{fig:skeleton}
\end{figure}

\section{EXPERIMENTS}

In this section, we investigate the performance of the proposed temporal CVAE model. Our aim is to anticipate human motion in an online setting and to use these predictions to infer the human's intention. For HRI, the understanding of the future human actions is of advantage for safe and task oriented action planning by the robot.

\begin{figure*}[th!]
\centering
\includegraphics [ width= 0.99 \textwidth]{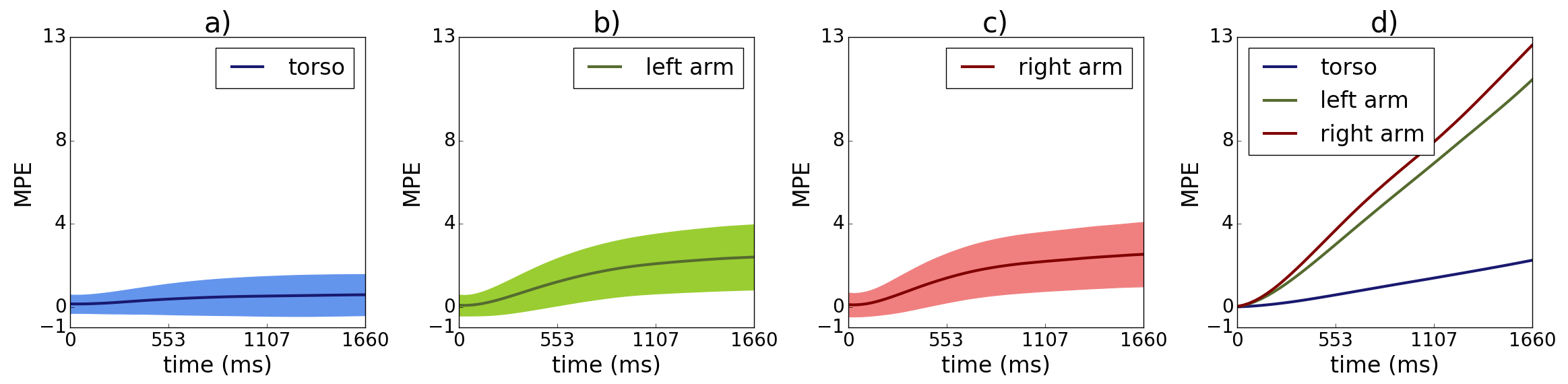}
\caption{The motion prediction error computed on the test set.  The mean and variance of the prediction error are presented for the CVAE's trained for the a) torso, b) left arm and c) right arm. The prediction error made by the linear model d) is depicted for all three limbs separately.} \label{Fig:mpe} \vspace{-0.3cm}
\end{figure*} 

\subsection{Data collection and model specifications}
\label{sec:models}

We collected skeletal data of the upper body with a Kinect sensor at 30 fps \cite{zhang2012microsoft}. The Kinect reference frame is centered at the Kinect, with the y-axis pointing upwards and the z-axis pointing towards the user. The upper body consists of nine joints as indicated and enumerated by Fig. \ref{fig:skeleton}. In total, we collected 90 minutes of random movements. The joints $I_j = [1:8]$ in each frame $\mathbf{f}_{t,I_j}$ were translated into a local reference frame which was centered at the root joint $\mathbf{f}_{t,[0]}$. The index $[...]$ indicates the index locations of the joints as enumerated in Fig. \ref{fig:skeleton}. Furthermore, we normalize the limb length to be of length one. This data set was further processed into input and output data by concatenating frames of each recording $r$  into sample pairs $\{ \mathbf{F}_{t -\Delta t}, \mathbf{F}_{t +\Delta t} \}_ {t\in [\Delta t:T_r-\Delta t]}$  where $T_r$ is the length of recording $r$. We set $\Delta t = 50$, i.e. that each time window covers approximately 1660 ms. 

As the structure of the human body is hierarchical, we use this prior information and train four separate models. First of all, in order to convert a prediction back into the original context, we require a model that predicts the translation of the root joint for every time step. Thus, the first model is trained on the input-output pairs \{$\mathbf{F}_{t-\Delta t,[0]}$, $\mathbf{F}_{t +\Delta t,[0]}$\}$ $.  Furthermore, we train separate models for the main body, left arm and right arm. For this, we define the index sets $I_{torso} = [1,2,3,4]$, $I_{right} = [3,5,7]$ and $I_{left} = [4,6,8]$, resulting in the training pairs \{$\mathbf{F}_{t-\Delta t,I_{torso}}$, $\mathbf{F}_{t +\Delta t,I_{torso}}$\}, \{$\mathbf{F}_{t-\Delta t,I_{right}}$, $\mathbf{F}_{t +\Delta t,I_{right}}$\}  and \{$\mathbf{F}_{t-\Delta t,I_{left}}$, $\mathbf{F}_{t +\Delta t,I_{left}}$\} respectively for all time steps $t$. For simplicity, let $limb \in [torso, right, left]$  denote any of these limbs. Each of these models is equipped with a parameter set $\theta_{limb} = [\theta^e_{limb}, \theta^t_{limb}, \theta^d_{limb}]$ and we denote the set of all parameters by $\lambda = [\theta_{torso}, \theta_{right}, \theta_{left}]$.

As depicted on the right in Fig. \ref{fig:generativemodel}, all  encoders consist of one layer of 200 units and 20 hidden variables $\mathbf{z_{t-1}}$, the transitioners consist of one layer of 30 units and 20 hidden variables $\mathbf{z_{t}}$ and the decoders consist of one layer of 200 units and the output dimension. The hidden variables and the output of the decoder require estimates of both a mean and a covariance for each dimension.  We train these models with the Adam optimizer  \cite{kingma2014adam} using a learning rate of $0.001$ and mini-batches of size 1000. The models are trained until the log-likelihood converges.
 
For comparison in the following sections we are making use of a linear model. Under the assumption of constant velocity this model predicts future time steps by estimating the current velocity $\mathbf{v}$ as the average velocity during the last 20 frames. The velocity  $\mathbf{v}$ is of dimension $(|I_{limb}|,3)$, where $|w|$ is the size of set $w$. A prediction of $\tau$ steps into the future at time $t$ is given by $\mathbf{\hat{f}}_{t+\tau,I_{limb}} = \mathbf{f}_{t,I_{limb}} + \tau \mathbf{v}$.

\subsection{Online human motion prediction}
\label{sec:onlinepred}

\begin{figure}[b!]
\centering
\includegraphics [ width= 0.44\textwidth]{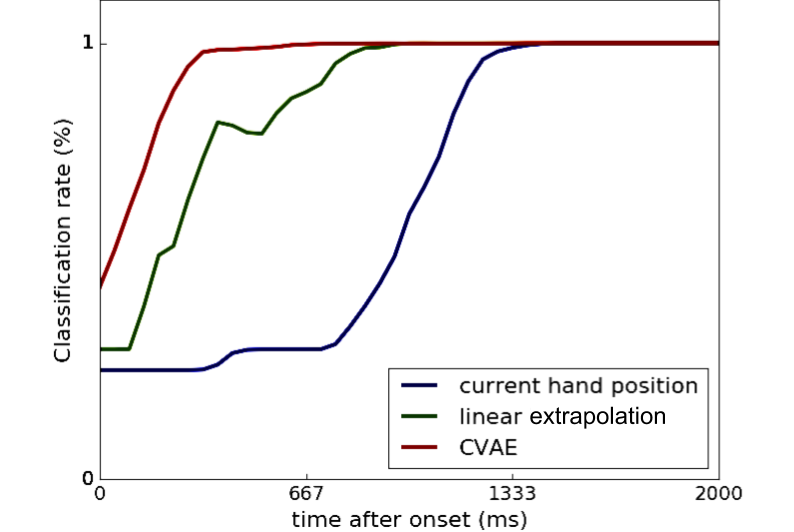}
\caption{End-point classification based on four different target locations. We compare the performance of the current hand position, a linear extrapolation and the CVAE model over time after movement onset. } \label{Fig:cla}
\end{figure}

\begin{figure*}[t!]
\centering
\includegraphics [ width= 0.9\textwidth]{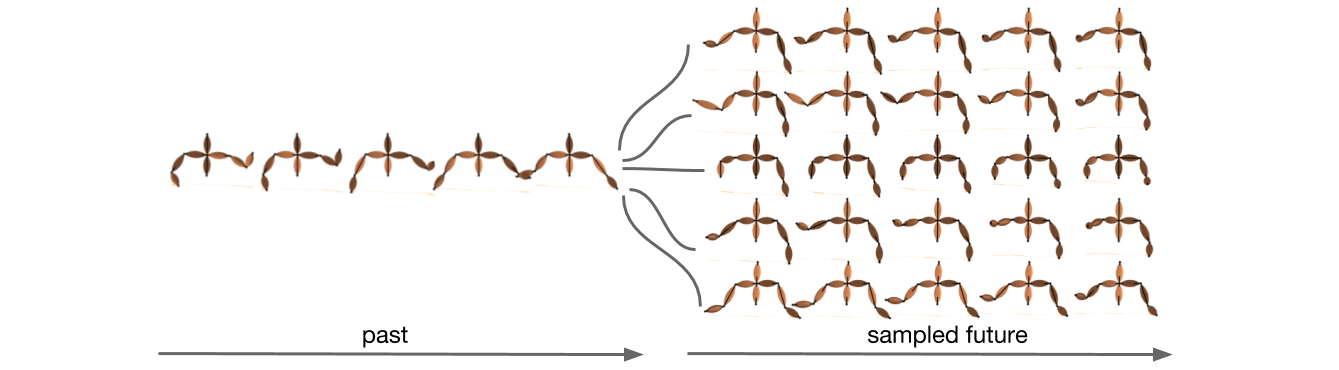}
\caption{Five different samples of possible futures. The samples were generated by propagating the past motion window through the network, sampling from the encoder and transitioner distributions and visualizing the mean output of the decoder. We depict the past 800 ms and samples of the next 800 ms. } \label{Fig:sam} \vspace{-0.3cm}
\end{figure*}

\begin{figure}[b!]
\captionof{table}{End-point classification rate} \label{tab:cla}
\resizebox{1\linewidth}{!}{
  \begin{tabular}{|l|c|c|c|c|}
    \hline 
     
    \multicolumn{1}{|l|}{Method} & \multicolumn{4}{|c|}{Classification rate  }    \\ \hline   
    \hline  
    \multicolumn{1}{|c|}{\% of traj.} &  20 &   43 &  60&   80  \\ \hline \hline
     \multicolumn{1}{|c|}{PFT \cite{perez2015fast}} & 73.26  & 89.55  & 95.76& 98.06   \\ \hline 
      \multicolumn{1}{|c|}{GMM \cite{perez2015fast}} & 57.08  & 85.83  & 96.94& 99.24   \\ \hline 
      \multicolumn{1}{|c|}{lin. extr. } & 53.47  & 79.25  & 95.4& 99.91   \\ \hline
     \multicolumn{1}{|c|}{CVAE } & \textbf{89.17}  & \textbf{98.93}  & \textbf{99.89}& \textbf{99.97}   \\ \hline
  \end{tabular}
  }
\end{figure}

As our models are trained to predict future human motion, we present the predictive performance for a held out test data set containing 20 minutes of arbitrary upper body movements. All errors are computed based on normalized data samples. Let $T_{test}$ be the number of frames of the test data. For each time $t \in [\Delta t:T_{test}-\Delta t]$ and each limb index set $I_{limb}$, we predict the next frames $\mathbf{F}_{t +\Delta t,I_{limb}}$ based on the past frames $\mathbf{F}_{t -\Delta t,I_{limb}}$. For every time step $t$ the decoder returns  mean estimates $\mu^d_{t,I_{limb}}$ and  variance estimates $\Sigma^d_{t,I_{limb}}$. We approximate the average motion prediction error (MPE) for each time step in the predicted time window $\Delta t$ by taking the mean squared error over all test samples and summing over the joint and spatial dimensions as follows
\begin{align}
MPE_{limb} = \sum_{j \in I_{limb}} \sum_{k \in [x,y,z]} \frac{1}{N} \sum_{t=\Delta t}^{T_{test}-\Delta t} (\mu^d_{t,I_{limb}} - \mathbf{F}_{t +\Delta t,I_{limb}})^2_{j,k},
\end{align}
where $N = T_{test}-2 \Delta t$.
Additionally, we report a variance estimate for each  time step in the predicted time window $\Delta t$  as the average sum of variances of the limb and spatial dimensions.
In Fig. \ref{Fig:mpe} a)--c) we visualize the motion prediction errors  of the torso, right arm and left arm model for the duration of 1660 ms. Since the skeleton is represented in a local reference frame, any natural movement of the torso is restricted to rotations. Therefore, the prediction error is comparatively low. The MPE for both arms is similar and grows more strongly than for the torso. Interestingly, the model seems to learn that there is less uncertainty for the initial position of the predictions. Note that due to the stochasticity in human motion, an accurate longterm prediction ($>$ 560 ms) is often not possible \cite{fragkiadaki2015recurrent}. 
For HRI it is important to represent these uncertainties about motion predictions such that the robot can take these into account during motion planning. 
In comparison to our CVAE models, a simple linear extrapolation in Fig. \ref{Fig:mpe} d) showcases the importance of modeling dynamics. While the initial error is comparable to the CVAE, the error increases rapidly after the initial frames. 

\subsection{Target prediction}
\label{sec:target}

As described in the related work, accurate online target prediction of human motion has been mostly addressed with GMMs in supervised settings. Although our approach is unsupervised, we investigate whether we can infer target locations based on predictions of our models. For this, we collected ten reaching trajectories of the right arm towards each of four different target locations. These locations were approximately 25-30 cm apart as in \cite{perez2015fast} while the initial position was held constant.  We classify the target location by using data and predictions of the right hand joint, i.e. $I_j = 7$. First, we benchmark our approach on two different methods. The classification is performed as described in Sec. \ref{subsec:method_goal} by replacing the sample frame by the corresponding signal and setting the covariance to be the identity matrix. 
\makeatletter
\def\BState{\State\hskip-\ALG@thistlm}
\makeatother

\begin{algorithm}
\caption{Sample future motion trajectory}\label{algor}
\begin{algorithmic}[1]
\Procedure{sample limb}{$\mathbf{F}_{t-\Delta t}, \theta_{limb}, I_{limb}$}
\State $ \textit {sample } \  \mathbf{z}_{t-1},\mathbf{z}_t \sim \mathcal{N}(\mathbf{0}, \mathbf{I}) $
\State $ \textit {compute } \mathbf{e}_t = f_e(\mathbf{z}_{t-1}, \mathbf{F}_{t-\Delta t, I_{limb}}, \theta^e_{limb})$
\State $ \textit {compute } \mathbf{t}_t = f_t(\mathbf{z}_{t}, \mathbf{e}_t, \theta^t_{limb})$
\State $ \textit {compute } \mu^d_{limb}, \Sigma^d_{limb} = f_d(  \mathbf{t}_t, \theta^d_{limb})$
\State return $\mu^d_{limb}$
\EndProcedure
\Procedure{sample}{$\mathbf{F}_{t-\Delta t}, \lambda, I_{torso}, I_{right}, I_{left}$}
\State $\hat{\mathbf{F}}_{t+\Delta t} \gets zeros(\Delta t, N_j,3)$
\State  For $I_{limb} \in [I_{torso}, I_{right}, I_{left}]$
\State $\quad \textit{sample } \mu^d_{limb} = \text{SAMPLE LIMB}(\mathbf{F}_{t-\Delta t}, \theta_{limb}, I_{limb})$
\State $\quad \textit{set } \ \ \quad \hat{\mathbf{F}}_{t+\Delta t,I_{limb}} = \mu^d_{limb} $ 
\State return $\hat{\mathbf{F}}_{t+\Delta t}$
\EndProcedure
\end{algorithmic}
\end{algorithm}

 
To verify that anticipation does improve classification in the early stages of a reaching movement, we classify the target by using the current last twenty frames at each time step.  As shown in Fig. \ref{Fig:cla}, this approach requires the hand to be at the target position, around 1300 ms, for a 99.9 \% correct classification rate. A simple linear extrapolation for twenty frames as described in Sec. \ref{sec:models}, reaches 99.9 \% within the first 600 ms after motion onset. Nevertheless, it appears that the dynamics learned by the CVAE are necessary for accurate prediction after around 300 ms. 

Furthermore, we compare our method to the trajectory alignment method with probabilistic flow tubes (PFT)  and GMMs presented in \cite{perez2015fast}. The emphasis in  \cite{perez2015fast} lies on fast target prediction based on the current hand trajectory and relies on training data from four targets and three initial positions to classify the resulting twelve classes. Given a general model of human motion, the task of end--point classification is independent of the initial position, which renders the functionality of our experiment and the experiments presented in \cite{perez2015fast} to be the same. The results are presented in Table \ref{tab:cla} as the classification rate based on different percentages of the trajectory from the initial position until the goal is reached (approx 1300 ms).  Our results indicate that anticipating future trajectories is of advantage as a simple linear extrapolation produces only slightly lower rates compared to GMMs \cite{perez2015fast}. The CVAE outperforms all methods and gives accurate classification rates within the first 200--300 ms after motion onset. Early inferences about the human's intention are required for the robot to plan ahead. As our model performs in real time, these predictions can be integrated into an online system. 

 \begin{figure}[t!]
 \vspace{0.2cm}
  \centering
  \includegraphics [ width= 0.45  \textwidth]{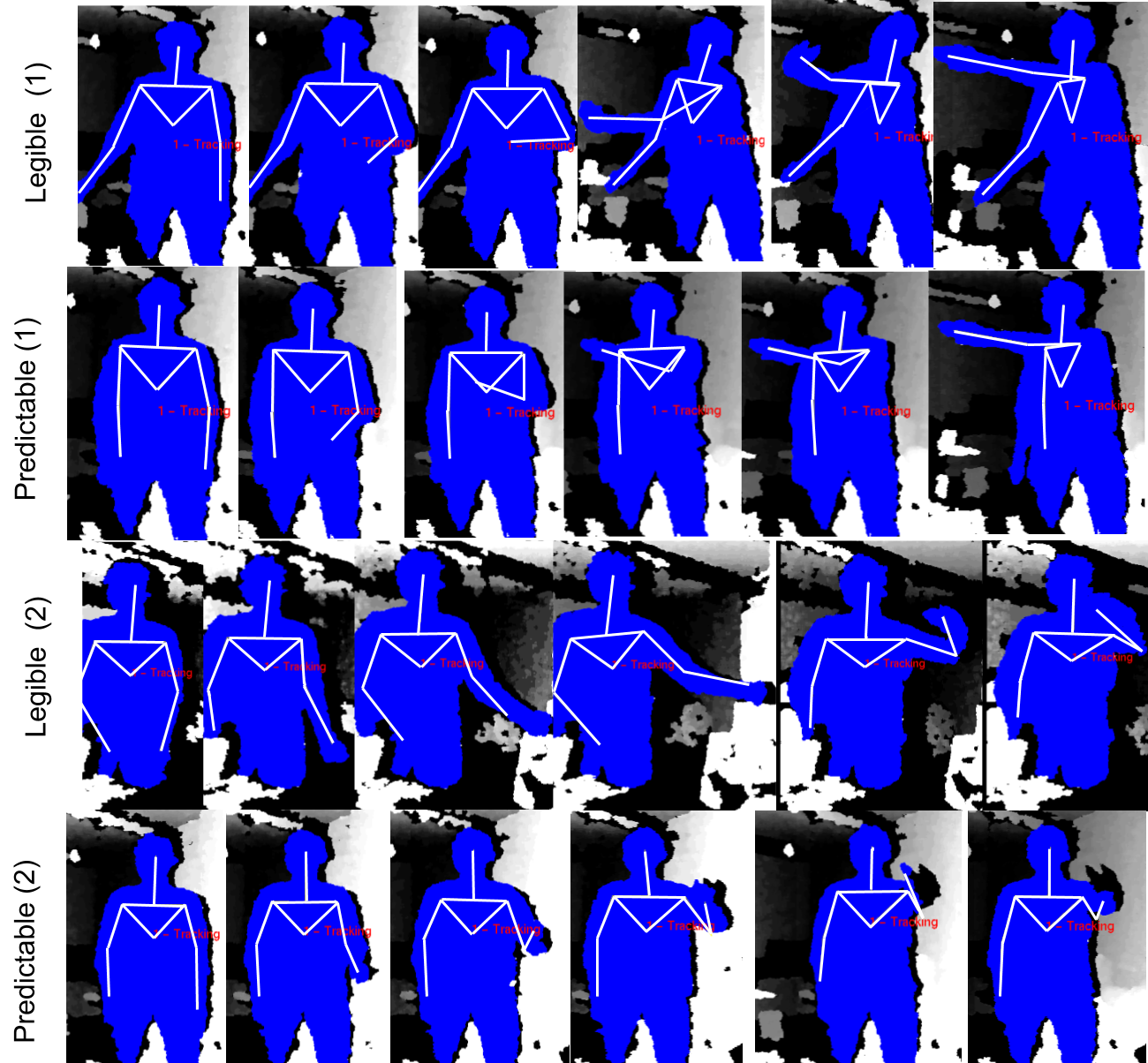}
  \caption{Examples of the legible and predictable motion patterns to target (1) and (2).} \label{Fig:legpred}
  \vspace{-0.3cm}
\end{figure}

\subsection{Sampling future motion trajectories}
\label{sec:sample}

The advantage of CVAEs over common autoencoders is that they model a probability distribution over future poses instead of a point estimate. Therefore, we can sample natural human motion from our model. In HRI it is of importance to be able to anticipate more than a single future. Even if the target of a human motion is fixed, the trajectories towards this goal can vary over time. 
The structure of the temporal CVAE allows us to sample from three different distributions, the encoder, the transitioner and the decoder. When propagating a window of past motion frames through the network at each of these layers, we can either use the mean estimate or a sample of the resulting distribution. When applying the mean in all layers, the output of the CVAE is a prediction. When sampling of at least one layer in the hierarchy, the output is a sample of the predictive distribution. To be of value for the robot, these samples should be in accordance with the current body configuration and represent natural motion patterns. 
In Fig. \ref{Fig:sam} we showcase five samples of future motion given a past sequence. As described in Alg. \ref{algor}, these samples were generated by propagating the input through the three models corresponding to the torso, the left arm and the right arm and sampling at both the encoder and transitioner stage. The resulting samples are in accordance with the input data, while varying slightly from sample to sample.  Online demonstrations of this method can be found at \href{https://www.youtube.com/watch?v=L93X3zh1sQo}{https://www.youtube.com/watch?v=L93X3zh1sQo}.

\begin{figure}[t!]
  \centering
  \includegraphics [ width= 0.45 \textwidth]{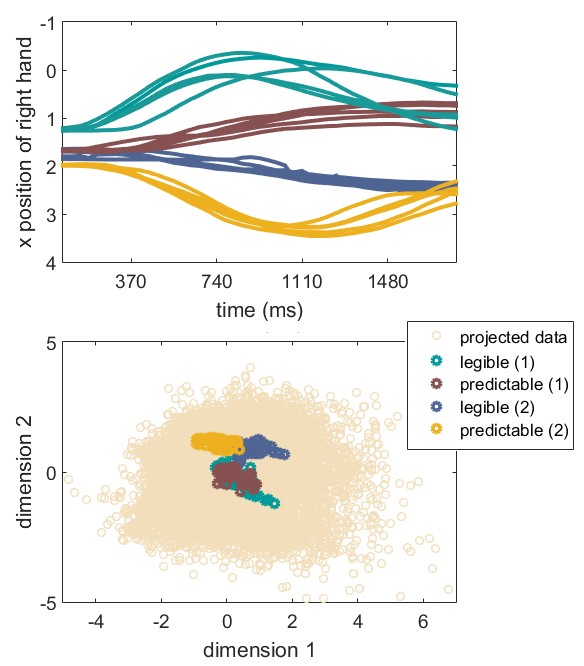}
  \caption{Top: The x position of the right hand during legible and predictable movements towards two targets. Lower: PCA of the activity of neurons in the inner layer of the model corresponding to the right arm. The majority of dots represents random test samples. The colored dots visualize the identity of testing samples corresponding to the reaching trajectories.} \label{Fig:rep} \vspace{-0.3cm}
\end{figure}

 \begin{figure}[b!]
\captionof{table}{Predictability vs. Legibility} \label{tab:pl}
\resizebox{1\linewidth}{!}{
  \begin{tabular}{|l|c|c|c|c|c|}
    \hline 
     
    \multicolumn{1}{|l|}{Traj. type} & \multicolumn{5}{|c|}{Classification rate    }    \\ \hline   
    \hline  
    \multicolumn{1}{|c|}{\% of traj.} &  20 &   43 &  60&   80 &   100\\ \hline \hline
    \multicolumn{1}{|c|}{legible (1)} & 73.89  & 99.99  & 100& 100   & 100   \\ \hline
      \multicolumn{1}{|c|}{predictable (1)} & 18.13  & 63.69  & 93.78& 97.02   & 100  \\ \hline
      \multicolumn{1}{|c|}{legible (2)} & 100  & 100  & 100& 100   & 100   \\ \hline
      \multicolumn{1}{|c|}{predictable (2)} & 95.79  & 89.96  & 94.7& 97.9   & 100  \\ \hline
      
  \end{tabular}
  }
\end{figure}

\subsection{Disentangling  legibility and predictability of human motion}
\label{sec:disent}

Due to the encoding--decoding structure of temporal CVAEs, these models do not only provide a mechanism to predict and generate natural human motion but they also learn a low dimensional manifold of the data. Ideally, this manifold should disentangle factors of variation. For example, when moving an arm along a line, the underlying dynamics could be explained by a single dimension. When the body rotates and the arm movement is kept along the same line, the first dimension must not change, but an additional dimension needs to account for the rotation. Predictable and legible motion of a reaching movement as introduced in Sec. \ref{sec:intro} should live in the same low dimensional space but be distinguishable within this space. A well separated low dimensional space would facilitate the robot's decision making compared to noisy signals of high dimensional arm trajectories. 

In this experiment, we investigate whether these theoretical assumptions hold true for the manifold learned by our model. We recorded legible and predictable reaching movements towards two targets, (1) and (2), with five recordings each, as exemplified in Fig. \ref{Fig:legpred}. The targets were positioned an arm length from the left (1) and the right (2) shoulder. In the top of Fig. \ref{Fig:rep} we depict the position of the right hand for all trajectories. Note, that we base all results on whole arm trajectories over time which corresponds to 450 dimensions for a single time window. To visualize the learned representation, we apply principal component analysis to the output of the right arm encoder layer of all testing samples. In the bottom of Fig. \ref{Fig:rep} we show the projection of 2000 samples onto the two main components. Furthermore, we project the neural output corresponding to the legible and predictable motion trajectories in different colors. 
It becomes apparent that the movement  types towards goal (2) are well separated while the movements towards target (1) are overlapping. Intuitively, the legible movement towards (2) reaches rather natural away from the body and towards the target, see third row of Fig. \ref{Fig:legpred}. Thus, both the hand and the elbow follow a different trajectory than during a straight reach as shown in the forth row of Fig. \ref{Fig:legpred}. Since target (1) lies to the left, the legible movements require an unnatural turn of the arm along the body, see first row of Fig. \ref{Fig:legpred}. As the elbow does not significantly deviate from the predictable trajectory, shown in the second row of Fig. \ref{Fig:legpred}, the two movements are less distinguishable in the latent space. 
An additional factor could be that the model has not learned to represent unnatural human motion as this is not contained in the training data. 

In order to verify whether legible motion has a positive impact on target inference, we classify the end-location of all movement types following the procedure described in Sec. \ref{subsec:method_goal}. As Table \ref{tab:pl} shows, legible motion indeed facilitates target inference for both targets. Especially during the initial phase of the movement, legibility seems to be of high importance.

\section{CONCLUSIONS AND FUTURE WORK}
\label{sec:conc}

In this work we presented a probabilistic deep learning approach to online human motion prediction for HRI. To overcome the drawbacks of using only the current body configurations for classification, we predict human motion for up to 1660 ms and base target inferences on these predictions. Our experiments suggest that our CVAE models are able to capture the underlying dynamics of general human motion. The Bayesian representation learning approach provides a method for sampling future trajectories and to investigate the underlying low dimensional structure of kinematic cues. Although this approach is data-intensive, the use of RGB depth images allows for low--cost recordings of natural, everyday movements. 

Instead of top--down approaches such as explicit cost functions or target specific training data, our approach is a bottom--up, data--driven model. The sensorimotor theory of predictive motion understanding in humans suggests that anticipation of sensory changes is important for online social interaction. We see our approach as a first step into this direction. In future work, we are aiming at integrating our system into online HRI tasks to investigate the importance of anticipating many futures in physical human--robot collaboration.

\addtolength{\textheight}{-12cm}   




\section*{ACKNOWLEDGMENT}
This work was supported by the EU through the project socSMCs (H2020-FETPROACT-2014) and the Swedish Research Council.

\bibliographystyle{IEEEtran}
\bibliography{root.bib}

\end{document}